\documentclass[a4paper, 10pt, conference]{ieeeconf} 
\IEEEoverridecommandlockouts
\usepackage{setspace}
\usepackage{graphicx}
\usepackage{cite}
\usepackage{amsmath}
\usepackage{color}

\usepackage[colorlinks=true]{hyperref}

\DeclareMathOperator*{\argmax}{arg\,max}

\title{
HeRoSwarm: Fully-Capable Miniature Swarm Robot Hardware Design With Open-Source ROS Support}
\author{Michael Starks, Aryan Gupta, Sanjay Sarma O V, and Ramviyas Parasuraman
\thanks{M. Starks and S. OV are with the College of Engineering, University of Georgia, Athens, GA, USA.}
\thanks{A. Gupta is with the School of Electrical and Computer Engineering, Georgia Institute of Technology, Atlanta, GA, USA.}
\thanks{R. Parasuraman is with the School of Computing, University of Georgia, Athens, GA 30602, USA.}
\thanks{Corresponding author email: {\tt ramviyas@uga.edu}}
}

\begin{document}

\maketitle

\begin{abstract}
Experiments using large numbers of miniature swarm robots are desirable to teach, study, and test multi-robot and swarm intelligence algorithms and their applications. To realize the full potential of a swarm robot, it should be capable of not only motion but also sensing, computing, communication, and power management modules with multiple options. Current swarm robot platforms developed for commercial and academic research purposes lack several of these critical attributes by focusing only on a few of these aspects. Therefore, in this paper, we propose the HeRoSwarm, a fully-capable swarm robot platform with open-source hardware and software support. The proposed robot hardware is a low-cost design with commercial off-the-shelf components that uniquely integrates multiple sensing, communication, and computing modalities with various power management capabilities into a tiny footprint. Moreover, our swarm robot with odometry capability with Robot Operating Systems (ROS) support is unique in its kind. This simple yet powerful swarm robot design has been extensively verified with different prototyping variants and multi-robot experimental demonstrations.
\end{abstract}

\section{Introduction}
A collection of independent miniature robots can be used as a sandbox for researching and testing swarm intelligence algorithms and multi-robot system applications such as pattern formation, self-organization, cooperative decision-making, etc. \cite{brambilla2013swarm,parker2007distributed,yang2020game}. 
Recently, modular swarm robots have utilized commercial off-the-shelf (COTS) hardware designs instead of custom designs that require special access or resource to produce a robot. 

The key design goals for the swarm robot hardware are to create a compact robot that includes an array of fundamental sensors, independent computing, precise self-localization, autonomous charging capability, and robust multi-level software control \cite{dorigo2021swarm,seeja2018survey}. 
In terms of software control and programming, it is ideal for making it compatible with the Robot Operating System (ROS) \cite{quigley2009ros}, which is a widely-used middleware framework in academic and commercial research robot platforms. This enables integration with any other robot platforms or simulators that are ROS-enabled. 
Having all these features in a single swarm robot platform would be highly desirable and valuable in analyzing and testing various sensing and control algorithms.

        \begin{figure}[t]
            \centering
            \includegraphics[width=0.98\columnwidth]{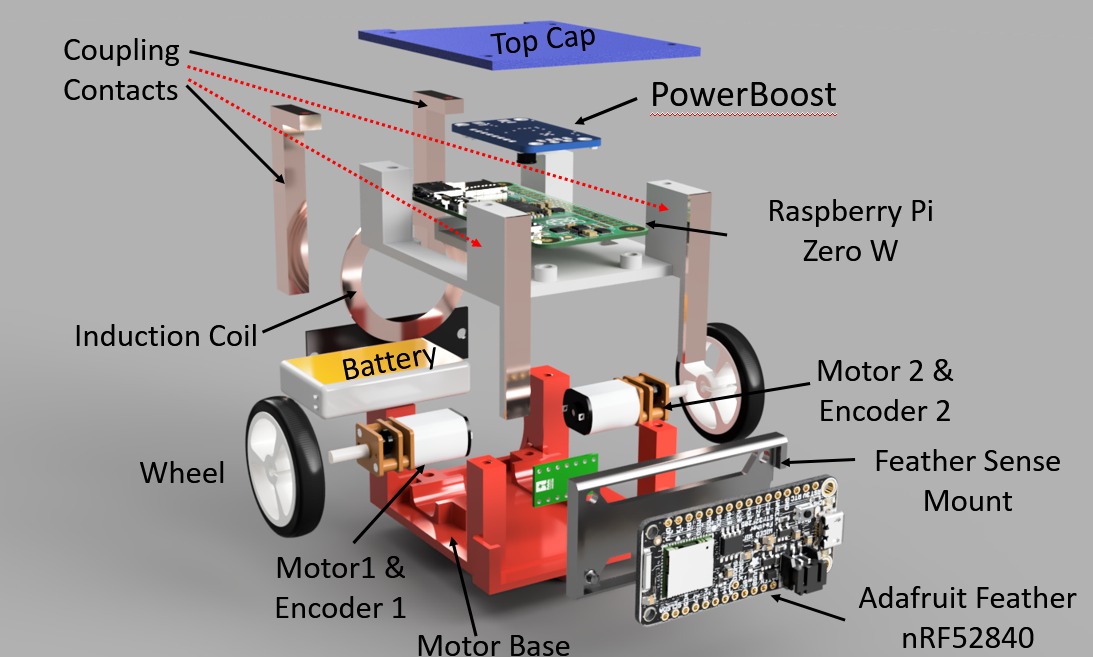}
            \caption{HeRoSwarm Robot Design - Exploded View Showing the Features.}
            \label{fig:swarm_robot}
        \end{figure}

Most of the current swarm robot platforms \cite{arvin2014development,cianci2006communication,khepera4,rubenstein2012kilobot,arvin2019mona,wilson2016pheeno} lack one or more of the outlined design goals and features.
For instance, the Robotarium multi-robot testbed \cite{wilson2021,pickem2015gritsbot} is a swarm algorithm simulator-hardware resource. The focus is on remote access and democratization of swarm robot experiments, while it lacks access to robot-level sensing data at the control interface. 
Among the ROS-supported platforms, Mona \cite{west2018ros}, WsBot \cite{wsbot} and the SMARTmBot \cite{jo2022smartmbot} stand out, but they either lack high-power computing or onboard odometry modules. 

To fill these gaps, in this paper, we propose the HeRoSwarm robot platform, which possesses the below unique features integrated with the ROS middleware framework. We open source the hardware design and software codes in GitHub\footnote{\url{https://github.com/herolab-uga/heroswarmv2}}. 
The key features and modules of the proposed HeRoSwarm design are as follows (see Fig.~\ref{fig:swarm_robot} for an overview of the design features):
\begin{itemize}
    \item \textbf{Sensing} Multiple sensors such as proximity, RGB, sound, Inertial (IMU) altitude, and humidity and pressure measurements to capture local information; 
    \item \textbf{Communication} Explicit data communication modalities such as Wi-Fi and Bluetooth; 
    \item \textbf{Computing} High C1-level \cite{trenkwalder2019computational} computing through a Raspberry Pi Zero-based computing module that can support ROS and advanced programming; 
    \item \textbf{Motion} Multi-level motor control with onboard wheel odometry aided by a microcontroller; 
    \item \textbf{Power} Dedicated power management with different recharging variants such as inductive wireless charging or magnetic coupling. 
\end{itemize}

The low-cost design uses COTS components (within a total cost of \$100) to ease replication by peer researchers. 
The HeRoSwarm platform will be a novel addition to the available swarm robot alternatives with its feature-rich design, compact size, and ROS compatibility. 
With these contributions and features, we expect that the HeRoSwarm robot will be applicable to wide use cases, including education and research on the basics of sensing, control, robotics, multi-robot systems, and swarm intelligence algorithms.

\section{Related Swarm Robot Platforms}

Many swarm robot designs were proposed in the literature (e.g., \cite{arvin2014development,choiRbot,rubenstein2012kilobot}), which primarily focused on specific applications or capabilities such as modularity for self-assembly \cite{wei2010sambot}, wheel-track mobility for rough terrains \cite{dorigo2013swarmanoid}, vibration-based drive for tiniest footprint \cite{kilobot}. Below, we analyze some closely relevant robot platforms used in the swarm robotics \cite{dorigo2021swarm} research domain.

\begin{figure}[t]
    \centering
    \includegraphics[width=0.55\linewidth]{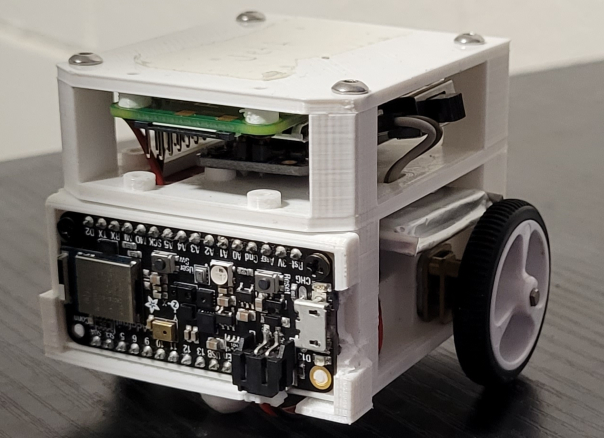}
    \includegraphics[width=0.41\linewidth]{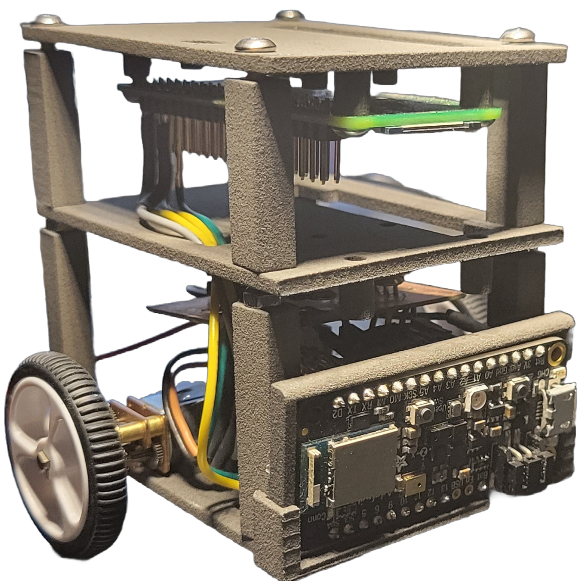}
    \vspace{-2mm}
    \caption{Sample HeRoSwarm prototypes used in validation experiments.}
    \label{fig:heroswarm-prototype}
    \vspace{-2mm}
\end{figure}
Georgia Tech's GRITSBot \cite{pickem2015gritsbot} is designed to be low-cost, compact robots that can be used to study swarm behavior. The GRITSBots' drive train is made of two stepper motors and subsequently forgoes wheel encoders and instead counts the change in the wheel's position through the stepper motor. There is not a readily available option to make the GRITSBot ROS enabled, and its onboard computing is a single microcontroller that is used to receive control commands and read the data from the 6 IR sensors that make up the sole sensor array on the robot. The GRTSBots have a way to charge autonomously with two extending contacts connecting to a metal charging strip.
    
The Kilobot \cite{rubenstein2012kilobot} is a markedly different swarm solution. These robots have the absolute minimum onboard computation necessary to carry out the programmed task; instead, a central computer controls all the robots in the swarm. The drive train for the Kilobot is made of vibration motors that allow to robot to slide across the work surface. The sensor included on the Kilobot allow for collision avoidance and ambient light sensing.
    
Khepera IV by Webots \cite{khepera4} is another research and educational robot platform. Each Khepera robot has a Linux core running on an ARM processor, and its sensor array includes eight infrared sensors for object detection, four line-following Infrared sensors, 5 Ultrasonic sensors for long-range object detection, a 3-axis accelerometer, a 3-axis gyroscope. The Khepera also has a microphone, a speaker, and a color camera. The capabilities of the Khepera are expandable through its KB-250 bus. The Khepera robots have a diameter of 140 mm and house a larger battery for approximately 7 hours of operation. By footprint, Khepera robots are larger in size than the HeRoSwarm Robots.
    
WSBots \cite{wsbot} are ROS-enabled miniature robots that can be used to model industrial environments. These robots are made to have capabilities that mimic tools used in an industrial environment. They feature wireless charging and lifts similar to that of a forklift. The WSbots lack extensive computing power due to the low-power computing unit used to achieve the small size and the absence of any environmental sensing. The WSBots were developed to explore a subset of problems in the swarm and multi-robot system space, limiting some capability to mimic the tools used in industry. 

A recent swarm robot design with a similar name HeRo2.0 \cite{rezeck2022hero}, has ROS compatibility and wheel encoders. While it has an array of proximity sensors, the robot is equipped with an ESP12 as the sole control unit to handle the motor control and network communication for the robot with the ROS middleware at a server. HeRo2.0 is a capable swarm research platform but lacks high-level compute capabilities comparable to HeRoSwarm, preventing resource-intensive expansion such as onboard computing with ROS.  
    
None of the existing swarm robots offer the sensing and computing capabilities that HeRoSwarm possesses in the same small form factor. 
They either possess good individual functionality or function as a multi-robot system, but the absence of an integrated solution can hinder their adoption. 
The available platforms do offer expansion options that are not possible with the current implementation of the swarm robot hardware. 
HeRoSwarm also differs from the other robot designs in terms of the integration of various critical hardware for a fully-capable swarm robot that can function both as a single robot and as well as a multi-robot platform. Open-source software support adds further novelty to HeRoSwarm design.
Fully functional prototypes of the HeRoSwarm robot designs are shown in Fig.\ref{fig:heroswarm-prototype}.

\section{HeRoSwarm Hardware Design}

\begin{figure}[t]
    \centering
    \includegraphics[height=0.6\linewidth]{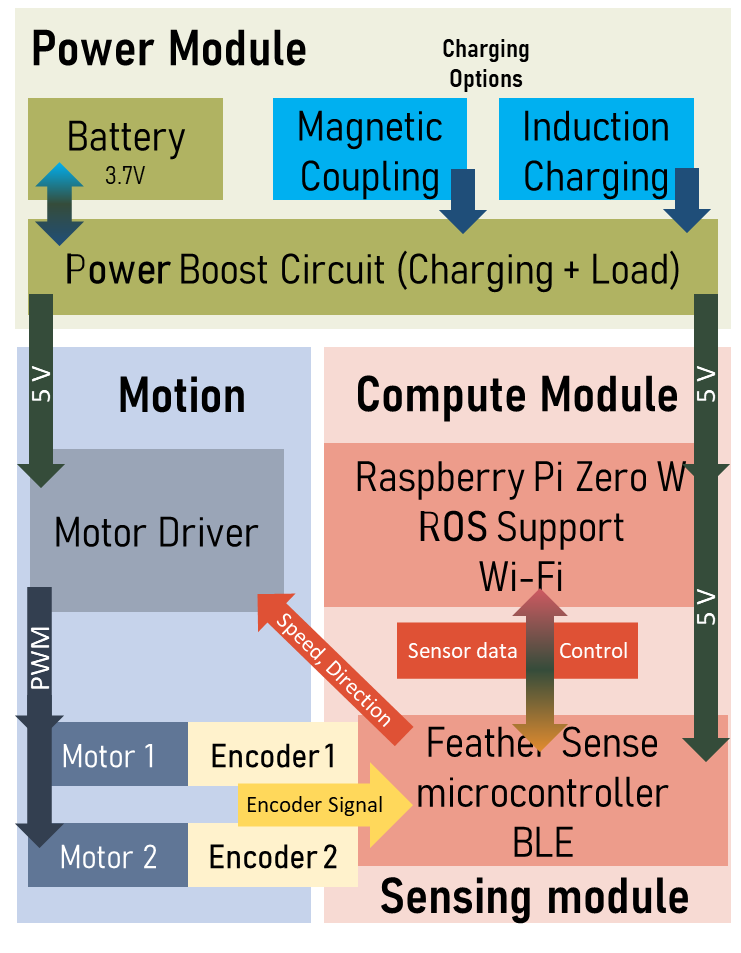}
    \includegraphics[height=0.59\linewidth]{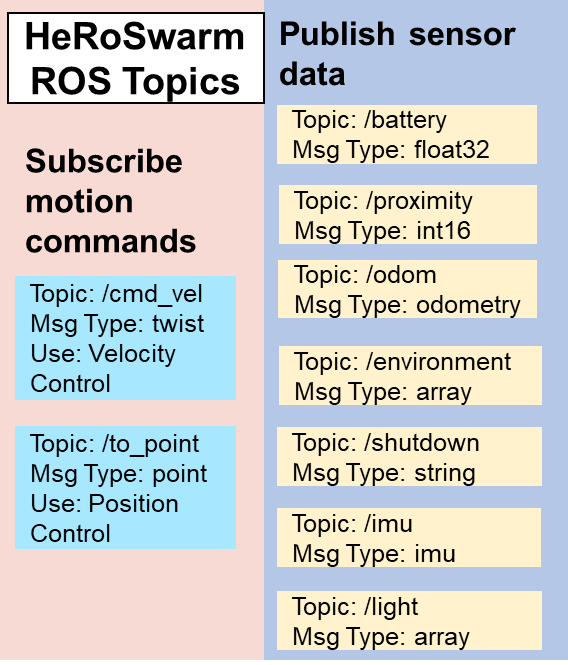}
    \vspace{-2mm}
    \caption{Hardware modules (left) and ROS software layout (right).}
    \label{fig:blockdia}
    \vspace{-2mm}
\end{figure}

\begin{figure}[t]
    \centering
    \includegraphics[width=0.8\linewidth]{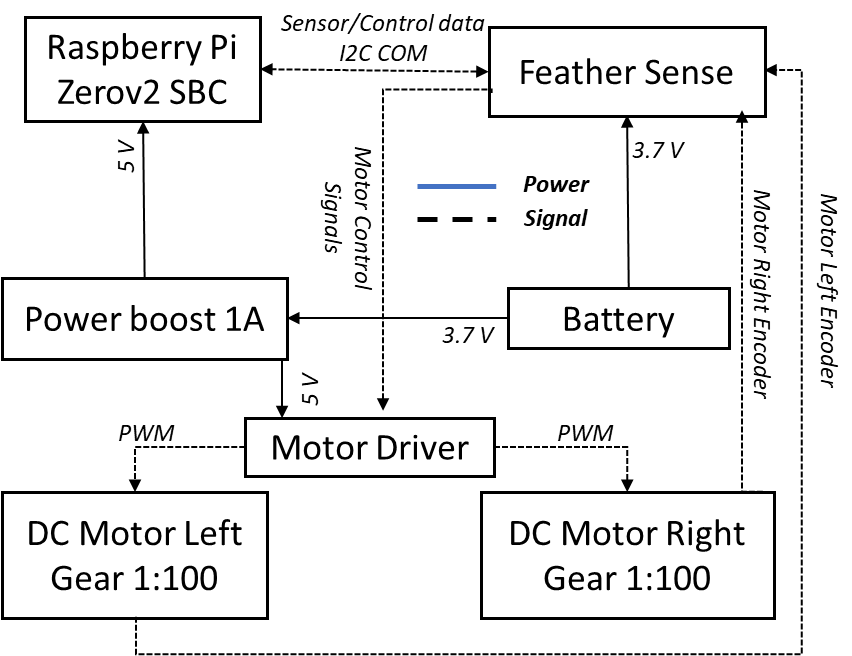}
    \vspace{-2mm}
    \caption{Schematic of the hardware connections in the HeRoSwarm robot.}
    \label{fig:connection-diag}
    \vspace{-2mm}
\end{figure}

With the HeRoSwarm (robot size 7cm x 7.3cm x 5.3 cm), users will have access to additional functionality, such as sensors and control granularity, while maintaining ease of use. Below, we describe different aspects of the robot design. See Fig.~\ref{fig:blockdia} for an overview of the integration of hardware modules as well as software elements. In Fig.~\ref{fig:connection-diag}, we present the connection schematic between the various modules.

\subsection{Sensing Component}
Unlike other swarm robot designs, HeRoSwarm is designed to have direct ROS-level access to a multitude of sensors and odometry through the integrated Feather Sense BLE board that houses various sensors along with an onboard microcontroller. 
The sensors included are the LSM6DS33 - 6 degrees of freedom accelerometer and Gyroscope that can determine the orientation of the robot; the LIS3MDL - a magnetometer that can also be used to determine the orientation of the robot; the APDS9960 - a light sensor that outputs color, proximity, and gesture data; and a BMP280 that measures temperature, altitude, and barometric pressure. The Sense board is also able to measure the humidity and process audio from a microphone. 
Additionally, each of the robot's wheels is equipped with a Hall Effect Quadrature Incremental Encoder that is connected to the Feather Sense.

\subsection{Computing and Communication}

The Broadcom BCM2835 SOC on the Raspberry Pi Zero runs a ROS-supported robot controller on top of an Ubuntu OS. This places the robot's design at the C1 computing level, as per the study in \cite{trenkwalder2019computational}.
While the Feather Sense is equipped with a Bluetooth module, the Raspberry Pi Zero W is equipped with an 802.11n Wi-Fi module. The HeRoSwarm robots are connected to the same wireless network that hosts the Central Server to send and receive information from the active ROS nodes. Communication between the Feather Sense and Pi can be achieved through I2C and UART ports. The addition of a Raspberry Pi module is optional, which is a feature of our modular design. Hence, the availability of ROS support is also based on the addition of a Pi module.

With ROS, the robot has a list of topics that exposes control of the robot and sensor data to the robot driver node (see Fig.~\ref{fig:blockdia}, Right). 
The robot's driver subscribes to a velocity topic that needs linear and angular velocity inputs, or the robot can be controlled through a position control node. The robot publishes all the sensor data on different topics at configurable publish rates depending on the application.

\subsection{Motion Control}

Each HeRoSwarm robot is equipped with two brushed DC motors used to control the position and orientation of the robots. The robots are powered through a boosted power to provide the 5 volts needed to power the Raspberry Pi and DC motors. 
The motors are controlled by two PID controllers on the sense board through a motor driver that sets the linear and angular speed of the robot based on the data from Pi zero. 
The robot follows a differential drive kinematics with linear and angular velocities calculated as 
\begin{equation}
\label{eq: robotLinearVelo}
    v_{R}=r.\frac{\phi_{right} +\phi_{left}}{2} ,  \; 
    \omega_{R}=r.\frac{\phi_{right}-\phi_{left}}{d_{w}},
\end{equation}
where $r$ is the radius of a wheel, $\phi$ is the wheel velocity applied to left or right motors, and $d_{w}$ is the distance between the centers of the wheels.

The velocities (inputs) received through the ROS framework are converted to independent wheel speeds by solving the inverse kinematics from Eq.~\eqref{eq: robotLinearVelo}. The velocities of the wheels are applied using a PID controller.

In addition, the built-in wheel encoders on the geared DC motors connected to the Feather Sense board are also used to track the robot's position (odometry) and velocity using the motor's encoders to know the robot's position in its coordinate frame. 
The odometry is updated every iteration of the Feather Sense main loop. 
The wheel encoders provide the number of ticks corresponding to the wheel rotations. The difference between the new and old wheel ticks is adjusted by varying the PWM frequencies. 
Here, an angle $\Delta \theta$ per tick can be computed from $\Delta T$ of ticks as
\begin{equation}
    \Delta \theta = A.\Delta T ,
\end{equation}
where A is the angle rotated per tick.
Thus the angular velocity of the motor,
\begin{equation}
\label{eq: differential}
  \omega = \frac{\Delta \theta}{\Delta t} = A.\frac{\Delta T}{\Delta t}
\end{equation}
and the corresponding linear velocity at each wheel is computed as 
\begin{equation}
       v = r.A.\frac{\Delta T}{\Delta t} = D_T \frac{\Delta T}{\Delta t}. 
\end{equation}
Here, the factor $D_T$ is the meters traveled by the wheel per tick which is pre-calibrated. 
The angle rotated by the robot in the $\Delta t$ interval is computed as $\Delta \Theta_{R}=\omega_{R} \Delta t$.
Also, the robot travels in a circle of radius $R$, computed as
\begin{equation}
    R=\frac{d_{w}}{2}\frac{(v_{right}+v_{left})}{(v_{right}-v_{left})} .
\end{equation}

The change in left and the right motor ticks are based on
\begin{equation}
\begin{split}
\Delta T_{right}&=\frac{2\frac{v_{R}}{D_T}+\frac{\omega_{R}}{A}d_{w}}{2R} \\
\Delta T_{left}&=\frac{2\frac{v_{R}}{D_T}-\frac{\omega_{R}}{A}d_{w}}{2R}\\
\end{split}
\end{equation}

Finally, the position update of the robot with respect to the global coordinate system in the HeRoSwarm testbed can be obtained from the following equations.
\begin{equation}
\begin{aligned}
    \label{eq: OdomTheta}
        X(t+1) &= X(t) + |v_{R}|.cos(\frac{\Delta \Theta_{R}}{2}).(2R sin(\frac{\Delta \Theta_{R}}{2})\\
        Y(t+1) &= Y(t) + |v_{R}|.sin(\frac{\Delta \Theta_{R}}{2}).(2R sin(\frac{\Delta \Theta_{R}}{2}) \\
        \Theta(t+1) &= \Theta(t) 
        +\frac{2D_T}{R}\Delta T
\end{aligned}
\end{equation}

\subsection{Power Management}

The robot consumes an average of 0.9W in idle mode and up to 1.5W when in motion. This brings the run time to 1-3 hours on average use with a 2000mAh battery capacity (LiPo 3.7V). The power is regulated using the Adafruit Powerboost 1000c. Each robot is equipped with a LiPo battery that powers the PowerBoost. The battery voltage is measured using the Feather Sense board and is used to track battery life. The battery data is monitored by the ROS node on the Raspberry Pi to activate the auto charge routine. The robots can be charged via multiple means: 1) using the micro-USB port available on the back of the Powerboost, 2) using the small copper plates with a magnetic coupling on the front and back of the robot, and (optionally) 3) using an induction coil to wirelessly charge the robot.

\section{HeRoSwarm Software Architecture}
\label{sec:software}
 The HeRoSwarm ROS workspace contains the Robot Controller, Camera Server, and Robot Messages packages needed for the multi-level controller. The purpose of the HeRoSwarm packages is to allow full user control of the swarm and the individual robots while being easy to use and integrate into current projects. The workspace uses standard topic names allowing integration with any ROS-enabled robot using standard topics and seamless integration of robots to form heterogeneous swarms \cite{parasuraman2020impact}. ROS allows for the use of the Publisher-Subscriber and Service-Client message architectures, facilitating communication between nodes on the same network. ROS also provides a way of distinguishing nodes with the same name through namespaces; appending the name of a robot to the node prevents duplication and erroneous control by distinguishing the nodes. The following sections detail the design for each package in the HeRoSwarm workspace and how they integrate to control the HeRoSwarm Robot platform.
    
The \textit{Robot controller} is a lightweight Python package that is deployed on the HeRoSwarm Robots. The controller exposes all of the features of the Robots to the ROS environment, including sensor data and odometry, in addition to position and velocity control. The controller is able to publish data from any of the sensors available on the Adafruit Feather sense board when the topics are activated. The odometry data gathered from the motor controller is published in the robot namespace to the Odom topic to allow users to read and track the robot's position in its frame. The Odometry data is stored in the ROS Odometry message format for ease of use with other ROS nodes. The position controller takes a tuple containing a position in the global frame that the robot should navigate to. The robot velocity can be controlled through the cmd\textunderscore vel topic; the linear velocity along the X-axis and rotational velocity along the Z-axis can be set independently for full control of each robot. The controller also creates topics for monitoring the battery level.

The \textit{Camera Server} is a key node in our HeRoSwarm ROS package. It provides tracking of the robot's position and orientation, the location of the charging stations, and creating the global position frame. The position of a robot is determined using an AprilTag \cite{olson2011apriltag} attached to the robot frame. The positions of all robots on the table are published on a global position topic. Each robot in the frame has a thread responsible for publishing its position data in the appropriate namespace and tracking the robot's odometry. The robot charging stations also use AprilTags to denote their position. The available charging stations are tracked by the camera server and assigned to the robot upon request from the low-battery service. The camera server is intended to run on a computer with a sufficient processing power for fiducial recognition, relieving some of the computational burdens from the swarm robots. Global position tracking is intended for evaluating the performance of swarm intelligence algorithms.

The multi-level control scheme gives the user freedom in deciding the granularity of their algorithm. Through the standard set of ROS Topics published by each robot, the user can monitor the state of the robot, set and get the robot's velocity and position, as well as have access to the onboard sensors. Each robot has its namespace, where it publishes and subscribes to a standardized list of topics to communicate with other controllers and robots. The Robot Controller includes a python class that neatly wraps the ROS functionality to facilitate swarm-level control. The wrapper class allows users to control the swarm as a collective rather than an individual robot. The number of robots used is determined by the user and can be changed asynchronously in the control loop; this allows users to add or remove robots at will to mimic any number of scenarios. The control loop steps at user-defined intervals to iterate through actions at different time scales. The wrapper can gather and return sensor data as well as the position data asynchronously of the control loop. The data is stored in a M x N matrix where N is the number of active robots and M is the tuple holding the data. The methods of the wrapper class can facilitate deployment onto the HeRoSwarm platform.

\section{Experimental Validation}

We made several prototypes of the HeRoSwarm robot design. Validation of the current design requires independent validation (unit testing) of all the implemented features before the whole system can be used. 
Below, we present the important results pertaining to the odometry and motion control, as they are unique to our robot.

\subsection{Localization}
As described in Sec.~\ref{sec:software} - Camera Server, a camera-based overhead position tracking system is used as ground truth to compare the robot's odometry and obtain state information of all robots in the system.

\begin{figure}
    \centering
    \includegraphics[width=0.45\textwidth]{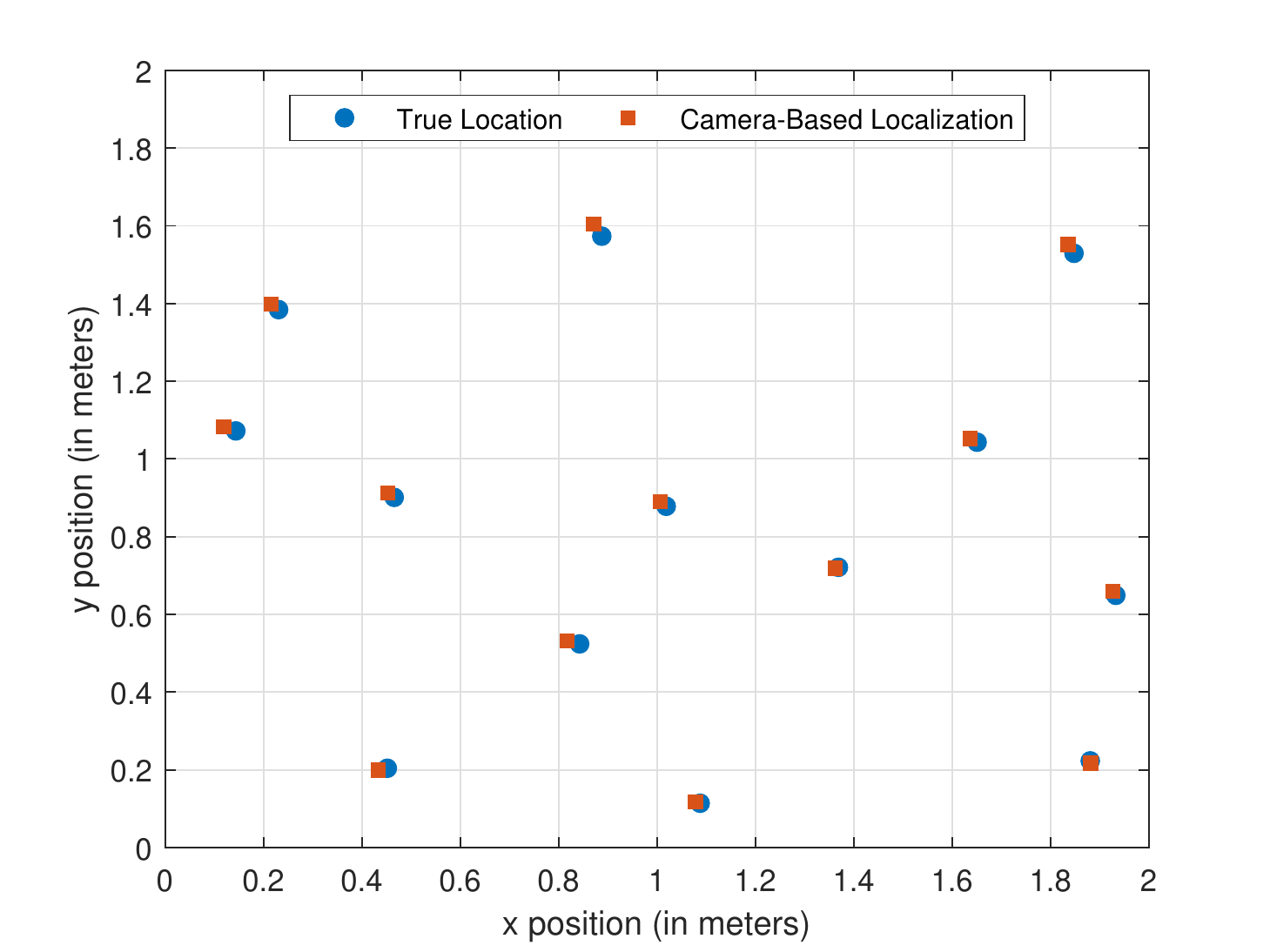}
    \vspace{-4mm}
    \caption{True Position Vs Camera-based Localization.}
    \label{fig:cam_pos_table}
\end{figure}

Each robot has a unique Apriltag that is used to locate its position in the global frame. The tag is pasted on the top plate of the robot chassis. Three tags are used to mark the axis of the operation field and are used to transform the robots into the global frame. Using Apriltags to mark the boundaries of the global frame allows for resizing the field to accommodate different experiments. The tabletop testbed hosting the swarm robot platforms has a size of 2.5 x 1.75 meters. To characterize the camera position system error, points were marked and measured on the board to find their position relative to the axis tags. An AprilTag was placed on each point, and the calculated position was recorded. Figure \ref{fig:cam_pos_table} shows the measured position and the position reported by the camera server. From the gathered data, the average error of the camera-based positioning system is 0.8 cm, meaning a reasonable sub-cm accuracy for using it as ground truth in robot tracking and performance evaluations.

\begin{figure}
    \centering
    \includegraphics[width=.45\textwidth]{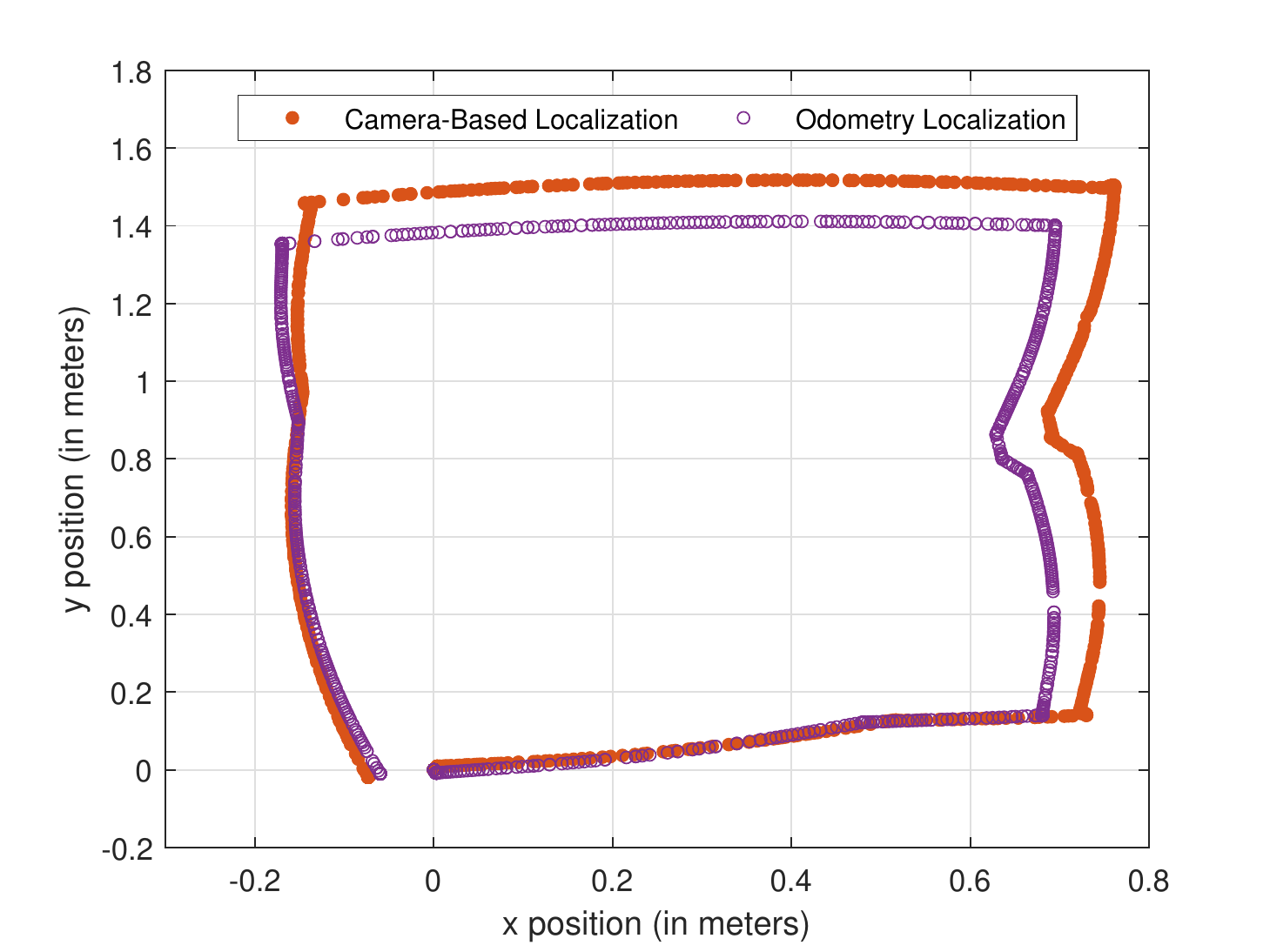}
        \vspace{-4mm}
    \caption{Camera-based Localization Vs. Odometry Localization.}
    \label{fig:odom_camera_error}
\end{figure}

The nRF52 on the Adafruit Feather Sense board calculates the robot's odometry. This information is passed to Raspberry Pi zero computer for publication on the Odom topic. To validate the accuracy of the odometry, the robot was driven along a defined path, and the camera position and the odometry position data were recorded. Figure \ref{fig:odom_camera_error} shows the position captured by the camera and the reported odometry. The camera position was shifted to originate at the starting position of the robot. The mean absolute error in odometry is 6.3 cm (with a standard deviation of 4.1 cm). The max error was observed to be 12.8 cm w.r.t. camera-based position. This shows the high odometry accuracy of the HeRoSwarm robots, which will play a crucial role in the evaluation of algorithms that do not rely on global positioning.
This accuracy can be further improved with the integration of odometry with IMU sensor data \cite{brossard2019learning}, which will be an avenue for future work.

\subsection{Motion Control}

Testing of the robot's motion control began by comparing the linear and angular velocities to the set velocities and finding the top speed. From the data gathered, the HeRoSwarm Robots have a max speed of 28 $cm \phantom{.} s^{-1}$. To ensure desired performance, the max linear and angular speeds are set such that when combined, the rotational speed of the wheels is below its maximum. We obtained a mean error of 4 cm/s in ten trials, showing promising results.

         \begin{figure*}
            \centering
            \includegraphics[width=0.30\linewidth]{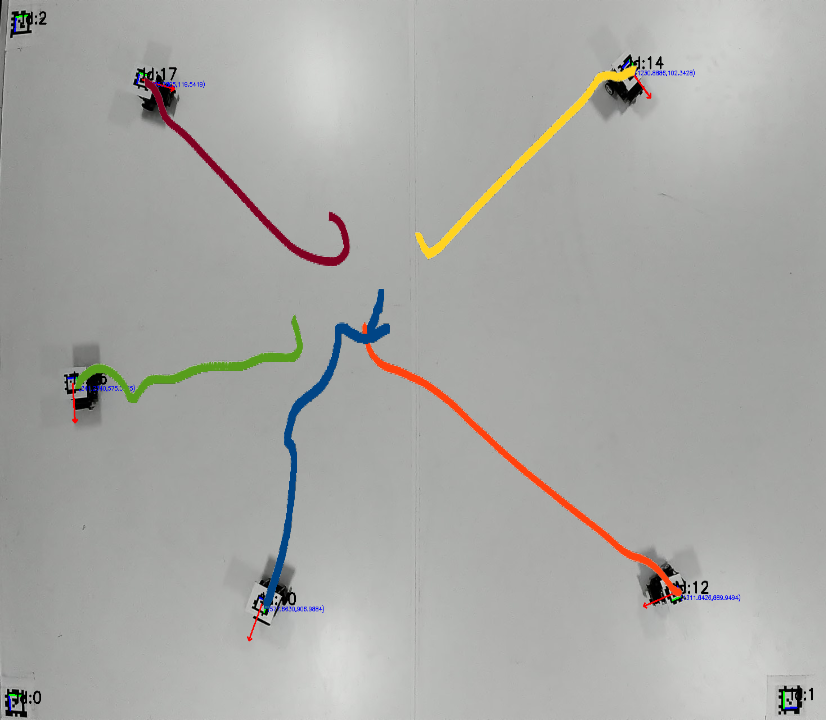}
            \includegraphics[width=0.30\linewidth]{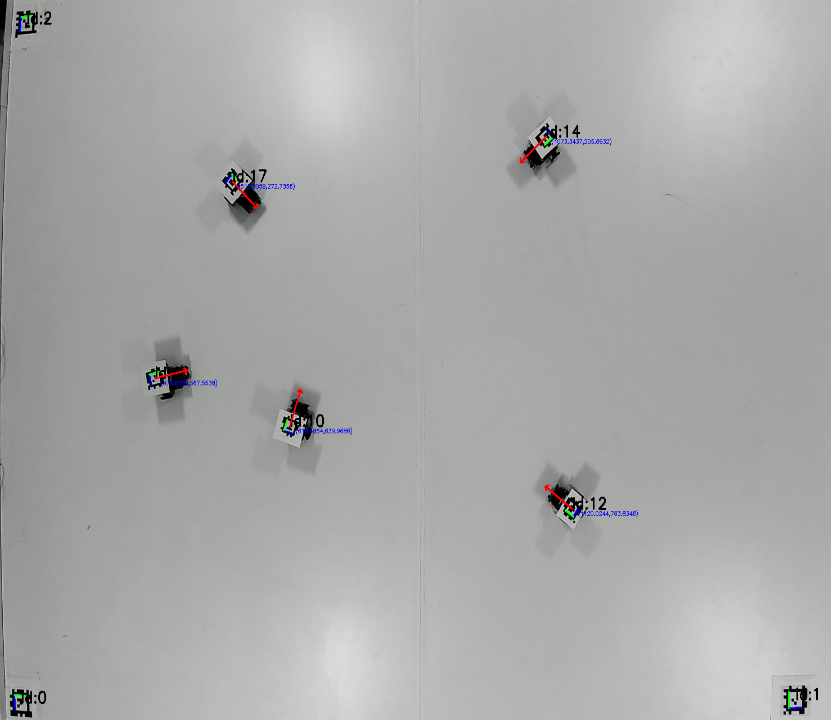}
            \includegraphics[width=0.30\linewidth]{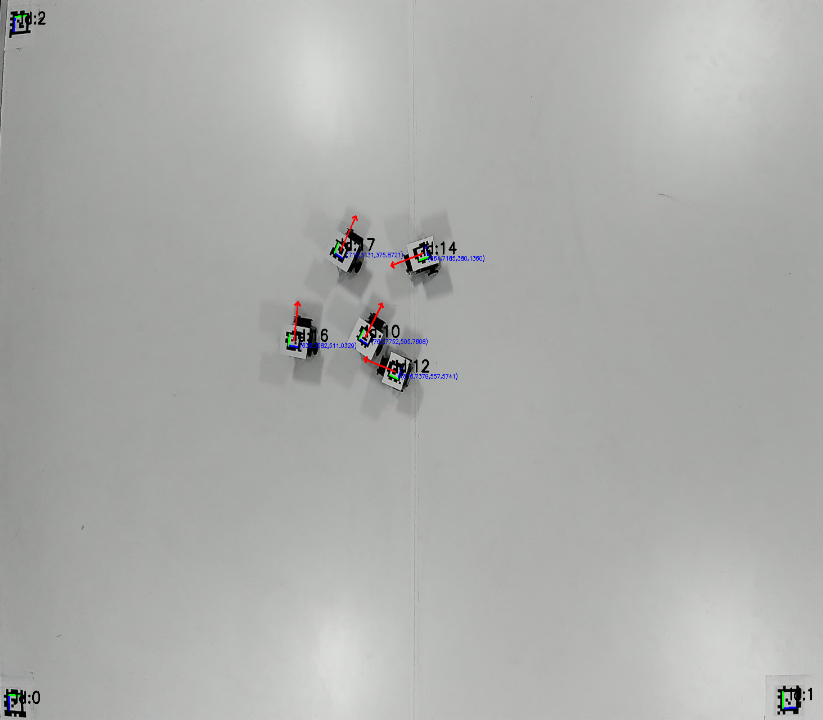}
                        \vspace{-2mm}
            \caption{Five swarm robots are executing the multi-robot rendezvous algorithm \cite{parasuraman2019consensus} (Sec.~\ref{sec:rendezvous}). Starting from a random initial position (left figure), their trajectory converges towards achieving closer proximity with each other, depicted by their intermediate (center figure) and the final instance (right figure).}
            \vspace{-2mm}
            \label{fig:swarm_rendezvous_3}
        \end{figure*}  
        \begin{figure*}
            \centering
            \includegraphics[width=0.30\linewidth]{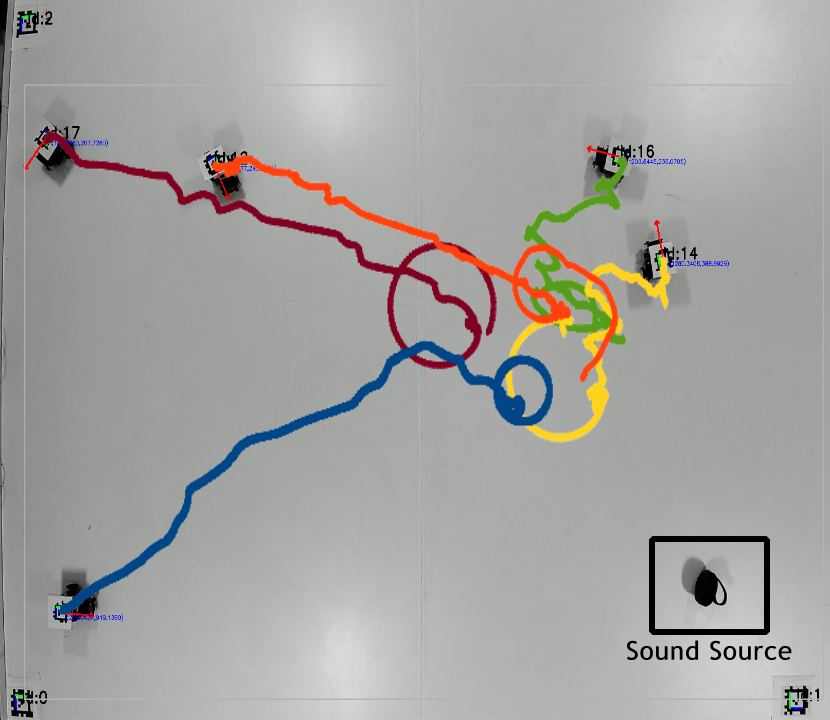}
            \includegraphics[width=0.30\linewidth]{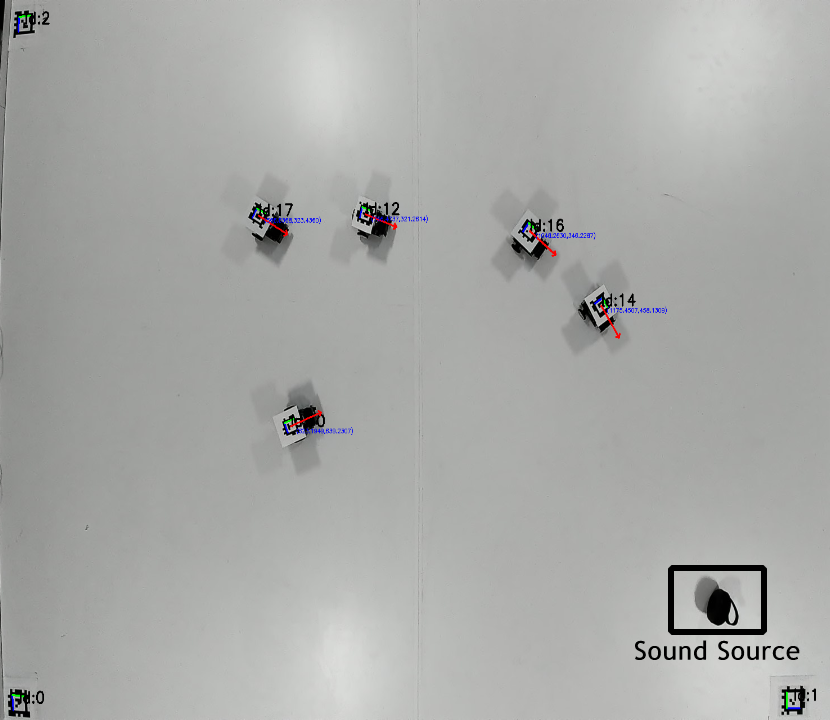}
            \includegraphics[width=0.30\linewidth]{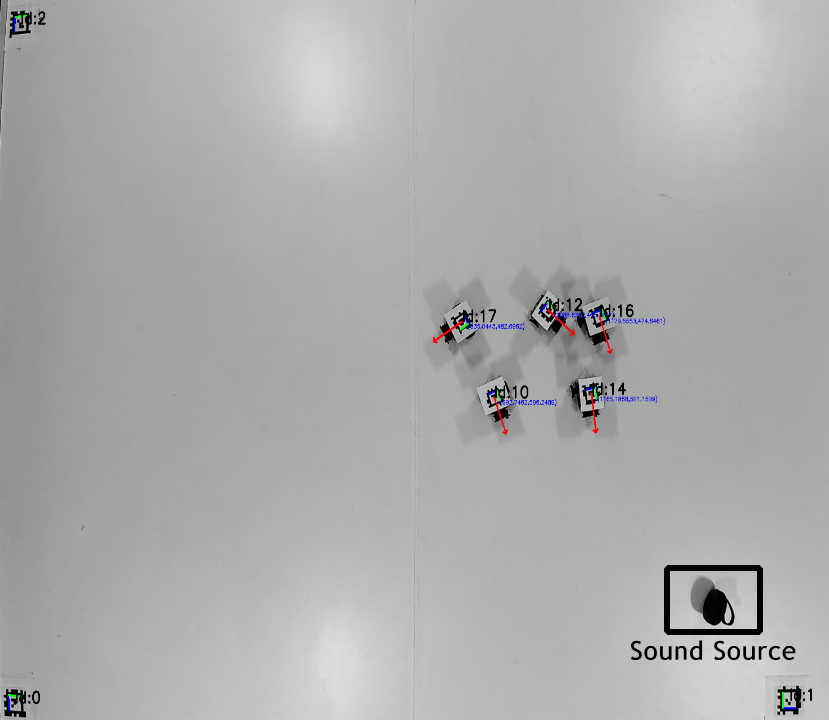}
                        \vspace{-2mm}
            \caption{Five swarm robots are executing microphone-assisted rendezvous (Sec.~\ref{sec:sound}). The neighbors of the robot that detects the loudest volume are converging on it. A Bluetooth speaker placed at the center-right is used to generate the sound source and the robots have reached close to this speaker.}
            \vspace{-2mm}
            \label{fig:swarm_sound_3}
        \end{figure*}  
          \begin{figure*}
            \centering
            \includegraphics[width=0.30\linewidth]{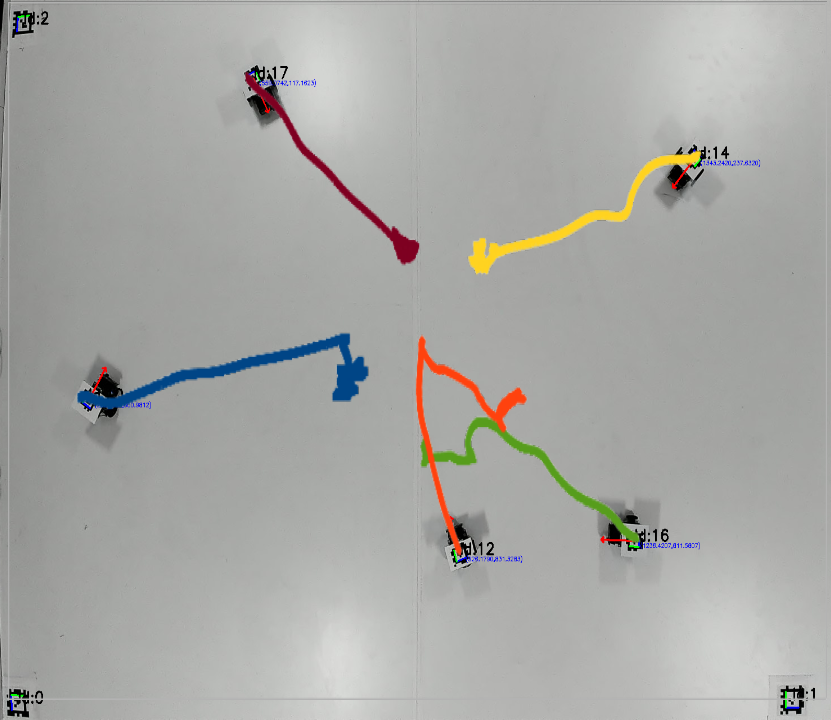}
            \includegraphics[width=0.30\linewidth]{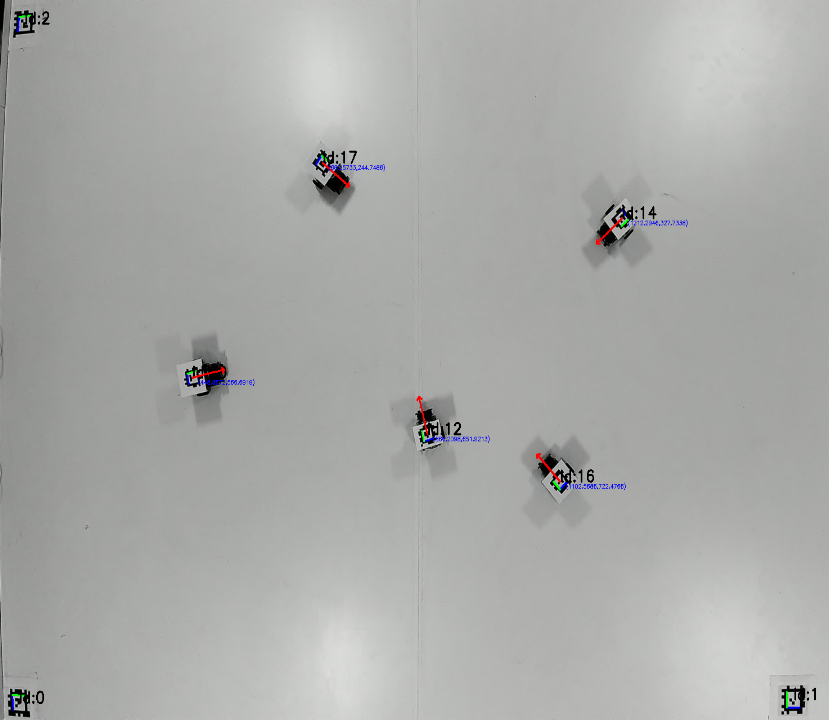}
            \includegraphics[width=0.30\linewidth]{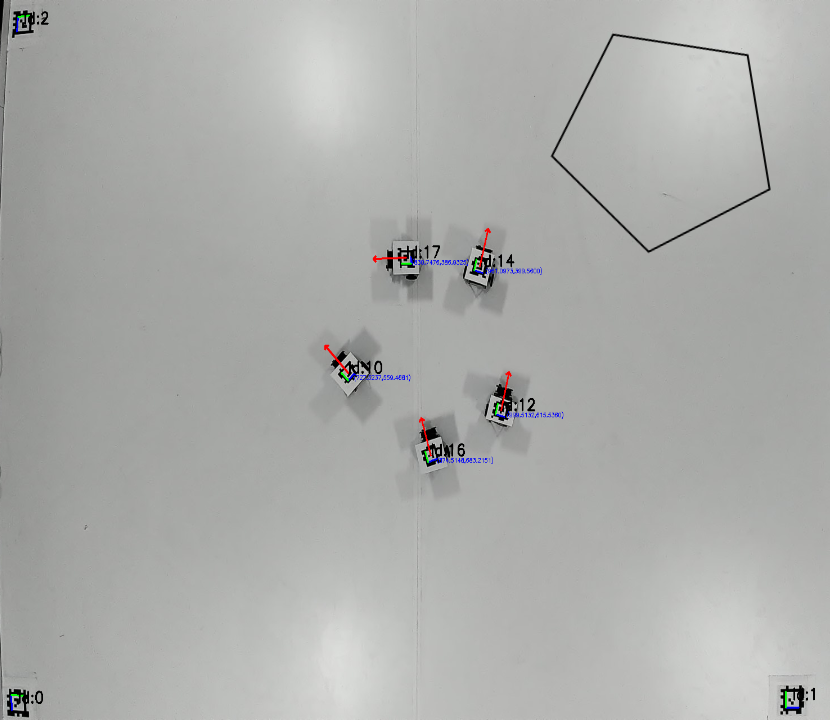}
                        \vspace{-2mm}
            \caption{Five swarm robots are executing the multi-robot formation control algorithm \cite{zavlanos2011graph} (Sec.~\ref{sec:formation}), starting from a random initial position and reaching a final stable position creating a pentagon-like predefined shape (depicted on the top-right corner of the right figure).}
            \label{fig:swarm_formation_3}
            \vspace{-2mm}
        \end{figure*}

\section{Demonstration with Multi-Robot Setup}
To holistically demonstrate the overall capability of the robot, we successfully implemented and tested three different multi-robot algorithms as detailed below. 

\subsection{Multi-Robot Rendezvous}
\label{sec:rendezvous}

We implement a typical rendezvous algorithm from \cite{parasuraman2019consensus}. The rendezvous problem here is to make the robots gather together quickly without a specific destination goal point, starting from their random initial positions. The rendezvous looks to control the robots such that the distance between the robots nears zero.
\begin{equation}
      \lim_{x \to \infty} (x_i - x_j) = 0, \forall i,j = 1 ... N
    \end{equation}
Here, $x_i$ is the position of robot $i$. From the initial state of the robots, the controller finds the central. The velocity ($\dot{x}$) with which each of the robots needs to move would be toward the center point of the robot's neighbors.
\begin{equation}
\dot{x}_{i} =\sum_{j \in N_{i}} (x_{j} - x_{i}),
\label{eqn:rendezvous}
\end{equation}
where $N_i$ is the neighboring robot to robot i. Each iteration of the algorithm further minimizes the distance between the robots, bringing them closer to the center.
    
The stop condition for the controller is bringing the robots as close together as possible. In the end, all the HeRoSwarm Robots will be as close together as the collision avoidance algorithm allows. Moving one of the robots will cause the others to reorient and minimize the position.
Fig.\ref{fig:swarm_rendezvous_3} shows the position (state) of the robots at different time instants during the rendezvous process, showing a successful demonstration of the rendezvous algorithm with the HeRoSwarm robots. 

\subsection{Multi-Robot Audio Signal-aided Rendezvous}
\label{sec:sound}
To further demonstrate the integration of sensors with motion control, we extend the rendezvous strategy in Eq.~\eqref{eqn:rendezvous} with sensor information. We use the microphone sensor on the Sense board to give us the intensity of the audio signal. 
In this demonstration, we follow the below equation.
\begin{eqnarray}
\dot{x}_{i} =\sum_{j \in N_{i}} (x_{max_j} - x_{i}), \\
max_j = \argmax_{j \in N_i} S_j .
\label{eqn:sound}
\end{eqnarray}
Here, $S_j$ is the sound intensity value of the microphone sensor on the robot, and $x_{max_j}$ is the robot's position that has the maximum audio signal intensity. So, all robots move toward the robot that is close to the audio source. 

The Lobar pickup pattern of the mic makes them highly direction-sensitive, meaning the robot closest to the mic does not necessarily detect the highest volume.
Fig.~\ref{fig:swarm_sound_3} shows the successful outcome of this demonstration, showcasing the utility of HeRoSwarm's sensor package and the sensor-motion integration in our ROS framework.

\subsection{Mutli-Robot Formation Control}
\label{sec:formation}
We implemented the muti-robot formation control algorithm from \cite{zavlanos2011graph}, which extends the rendezvous control algorithm by constraining their relative positions to lie within a formation shape. In this example, we used five robots that were commanded to group together and form a pentagon with a side length of roughly $.25$ meters.
Assume $\zeta_i$ defines a coordinate of robot $i$ within a specific formation shape containing all the robots. 
Then the formation control problem extends the rendezvous control in the below equation.

\begin{eqnarray}
\dot{x}_{i} =\sum_{j \in N_{i}} (\chi_{j} - \chi_{i}), \text{ where, } \chi_i = x_i - \xi_i .
\label{eqn:formation}
\end{eqnarray}
We tested this algorithm and the results are shown in Fig.~\ref{fig:swarm_formation_3}, demonstrating another strong integration of HeRoSwarm capability in tightly coordinated control tasks.

\section{Conclusion}
We proposed a new swarm robot platform called HeRoSwarm, which integrates full functional capability in terms of sensing, motion, computing, communication, and power management. The simple design with COTS components, robust sensor package (with a multitude of sensors), and onboard computation facilitate the implementation of complex and distributed algorithms. The robot's ROS-supported software architecture expands the accessibility and enables integration with the ROS software ecosystem. 
The unique onboard odometry gives real-time data on the position of the robots to different levels of control. 

The HeRoSwarm robot has a huge potential in swarm robotics and related applications, as its functionality exposed in the robot hardware and software design gives the user flexibility to implement any desired multi-robot and swarm intelligence algorithms (e.g., \cite{sarma2013development,latif2021energy,latif2022dgorl}). 
While the HeRoSwarm has met its primary design goals, it continues to undergo improvements to expose more features and additional hardware and software functionalities (compatibility with ROS2 for instance).
We validated the robot's odometry and motion control and demonstrated the sensor-motion and swarm system integration through multi-robot experiments.

\bibliographystyle{IEEEtran}
\bibliography{bib}
\end{document}